\documentclass[11pt]{article}
\pdfoutput=1
\usepackage[
  margin=0.83in,
  headheight=12pt,
  headsep=25pt,
  includefoot,
  footskip=30pt,
]{geometry}
\usepackage{tikz}
\usepackage{graphicx}
\usepackage{comment}
\usepackage{listings}
\usepackage{xcolor}
\usepackage{inconsolata}
\usepackage{makecell}
\usepackage{tabularx} 
\usepackage{adjustbox}
\usepackage{rotating}

\definecolor{commentcolour}{rgb}{0.3,0.7,0.2}
\definecolor{backcolour}{rgb}{0.98,0.98,0.98}

\lstdefinelanguage{markdown}{
    comment=[l]{\#},
    morestring=[s]{```}{```},
    commentstyle=\color{commentcolour}\bfseries,
    stringstyle=\color{blue},
    basicstyle=\scriptsize\ttfamily,
    showstringspaces=false,
    breaklines=true,
    breakautoindent=false,
    breakindent=0pt,
    backgroundcolor=\color{backcolour},
}
\lstdefinestyle{mystyle}{
    morekeywords={self},
    basicstyle=\scriptsize\ttfamily,
    keywordstyle=\color{blue},
    commentstyle=\color{commentcolour}\bfseries,
    breaklines=true,
    breakautoindent=false,
    showstringspaces=false,
    backgroundcolor=\color{backcolour},
    stringstyle=\color{red},
}
\lstdefinelanguage{PythonPlus}[]{Python}{
  alsoother={@},
  morekeywords=[1]{,as,assert,nonlocal,with,yield,self,True,False,None} 
  morekeywords=[2]{,__init__,__add__,__mul__,__div__,__sub__,__call__,__getitem__,__setitem__,__eq__,__ne__,__nonzero__,__rmul__,__radd__,__repr__,__str__,__get__,__truediv__,__pow__,__name__,__future__,__all__,}, 
  morekeywords=[3]{,object,type,isinstance,copy,deepcopy,zip,enumerate,reversed,list,set,len,dict,tuple,range,xrange,append,execfile,real,imag,reduce,str,repr,}, 
  morekeywords=[4]{,Exception,NameError,IndexError,SyntaxError,TypeError,ValueError,OverflowError,ZeroDivisionError,}, 
  morekeywords=[5]{,ode,fsolve,sqrt,exp,sin,cos,arctan,arctan2,arccos,pi, array,norm,solve,dot,arange,isscalar,max,sum,flatten,shape,reshape,find,any,all,abs,plot,linspace,legend,quad,polyval,polyfit,hstack,concatenate,vstack,column_stack,empty,zeros,ones,rand,vander,grid,pcolor,eig,eigs,eigvals,svd,qr,tan,det,logspace,roll,min,mean,cumsum,cumprod,diff,vectorize,lstsq,cla,eye,xlabel,ylabel,squeeze,}, 
}

\usepackage{tikz}

\usepackage{amsmath} 
\usepackage{array} 
\usetikzlibrary{matrix, arrows}
\usepackage{amsmath,amssymb}
\usepackage{amsthm}
\usepackage{mathtools}
\usepackage{xspace}
\usepackage[noend]{algorithmic}
\usepackage[ruled,vlined]{algorithm2e}
\usepackage{url}
\usepackage{makeidx}
\usepackage{enumerate}
\usepackage{epstopdf}
\usepackage{booktabs}
\usepackage{color}
\usepackage[utf8]{inputenc}
\usepackage{thm-restate}
\usepackage{scalerel,stackengine}
\usepackage[shortlabels]{enumitem}
\usepackage{xr}
\usepackage{fancyvrb}
\usepackage{xcolor}
\usepackage{bold-extra}
\usepackage{arydshln}
\usepackage[small]{caption}
\usepackage{subcaption}
\usepackage[most]{tcolorbox}
\usepackage{fvextra}
\usepackage{float}
\usepackage{alltt}
\usepackage{soul}
\usepackage{MnSymbol,wasysym}
\usepackage{fancyvrb}
\usepackage{multirow}
\usepackage[final]{hyperref}
\usepackage[bottom]{footmisc}
\usepackage{diagbox}
\usepackage{mdframed}
\usepackage{todonotes}
\setuptodonotes{inline}
\usepackage{tikz}
\usetikzlibrary{shapes,calc,positioning}

\global

\tcbset{
  aibox/.style={
    width=\textwidth,
    top=0pt, bottom=0pt, left=5pt, right=5pt,
    colback=white,
    colframe=black,
    colbacktitle=black,
    enhanced,
    center,
    attach boxed title to top left={yshift=-0.1in,xshift=0.15in},
    boxed title style={boxrule=0pt,colframe=white,},
  }
}
\newtcolorbox{AIbox}[2][]{aibox,title=#2,#1}

\definecolor{aigold}{RGB}{244,210, 1} 
\definecolor{aigreen}{RGB}{210,244,211} 

\sethlcolor{aigreen}

\definecolor{aired}{RGB}{255,180,181}

\newtcbox{\mybox}[1][green]{on line,
arc=0pt,outer arc=0pt,colback=#1!10!white,colframe=#1!50!black,
boxsep=0pt,left=0pt,right=0pt,top=0pt,bottom=0pt,
boxrule=0pt,bottomrule=0pt,toprule=0pt}

\newcolumntype{Y}{>{\centering\arraybackslash}X}

\title{Phi-3 Safety Post-Training: Aligning Language Models with a “Break-Fix” Cycle}
\author{Microsoft}
\date{}

\begin{document}

\maketitle

\begin{abstract}
Recent innovations in language model training have demonstrated that it is possible to create highly performant models that are small enough to run on a smartphone. As these models are deployed in an increasing number of domains, it is critical to ensure that they are aligned with human preferences and safety considerations. In this report, we present our methodology for safety aligning the Phi-3 series of language models. We utilized a “break-fix” cycle, performing multiple rounds of dataset curation, safety post-training, benchmarking, red teaming, and vulnerability identification to cover a variety of harm areas in both single and multi-turn scenarios. Our results indicate that this approach iteratively improved the performance of the Phi-3 models across a wide range of responsible AI benchmarks. Finally, we include additional red teaming strategies and evaluations that were used to test the safety behavior of Phi-3.5-mini and Phi-3.5-MoE, which were optimized for multilingual capabilities.
\end{abstract}

\section{Introduction}

Given the computational cost associated with large language models (LLMs), there is increasing interest in developing smaller models with lower compute and memory requirements. Recent research has demonstrated that it is possible to create performant small language models (SLMs) by training on highly curated and synthetic datasets \cite{gunasekar2023textbooks, li2023textbooks, javaheripi2023phi}. In April 2024, Microsoft released the Phi-3 series of SLMs, including Phi-3-mini (3.8B), Phi-3-small (7B), and Phi-3-medium (14B). For example, Phi-3-mini is small enough to run on a smartphone but achieves 69\% on MMLU and 8.38 on MT-Bench, making it competitive with Mixtral 8x7B and GPT-3.5. \cite{abdin2024phi3}. 

Open-source SLMs enable an exciting array of on-device generative AI applications. At the same time, the proliferation of language models in an increasing number of domains underscores the importance of aligning models to human preferences and safety considerations. In this report, we present our approach to aligning the Phi-3 series of language models. We utilized a “break-fix” cycle that relies on multiple rounds of vulnerability identification and safety post-training. In the sections that follow, we detail our methodology, quantitative benchmarks, and red teaming results. 

\section{Safety Alignment} 

\subsection{Approach}

Microsoft adopted an iterative approach to safety post-training that consisted of five main stages: 

\begin{enumerate}
\item \textbf{Safety Dataset Curation:} We utilized existing publicly available datasets with various modifications and generated additional datasets based on feedback from the AI Red Team for further safety post-training.
\item \textbf{Safety Post-Training:} The safety datasets mixed with standard preference datasets were used in both the supervised fine-tuning (SFT) and direct preference optimization (DPO) \cite{rafailov2023direct} stages.
\item \textbf{Quantitative and Qualitative RAI Evaluations:} A wide spectrum of responsible AI (RAI) evaluations, described in detail below, were considered to select release candidates (RCs) to share with the AI Red Team.
\item \textbf{AI Red Teaming:} The release candidates were shared with a centralized and independent AI Red Team (AIRT), which leveraged a variety of adversarial techniques to probe models for harmful content. The red teaming strategy is described below.
\item \textbf{Vulnerability Identification:} Based on the RAI evaluations and AIRT findings, potential vulnerabilities are identified to inform further safety post-training.
\end{enumerate}

\begin{figure}[h]
\includegraphics[scale=0.6]{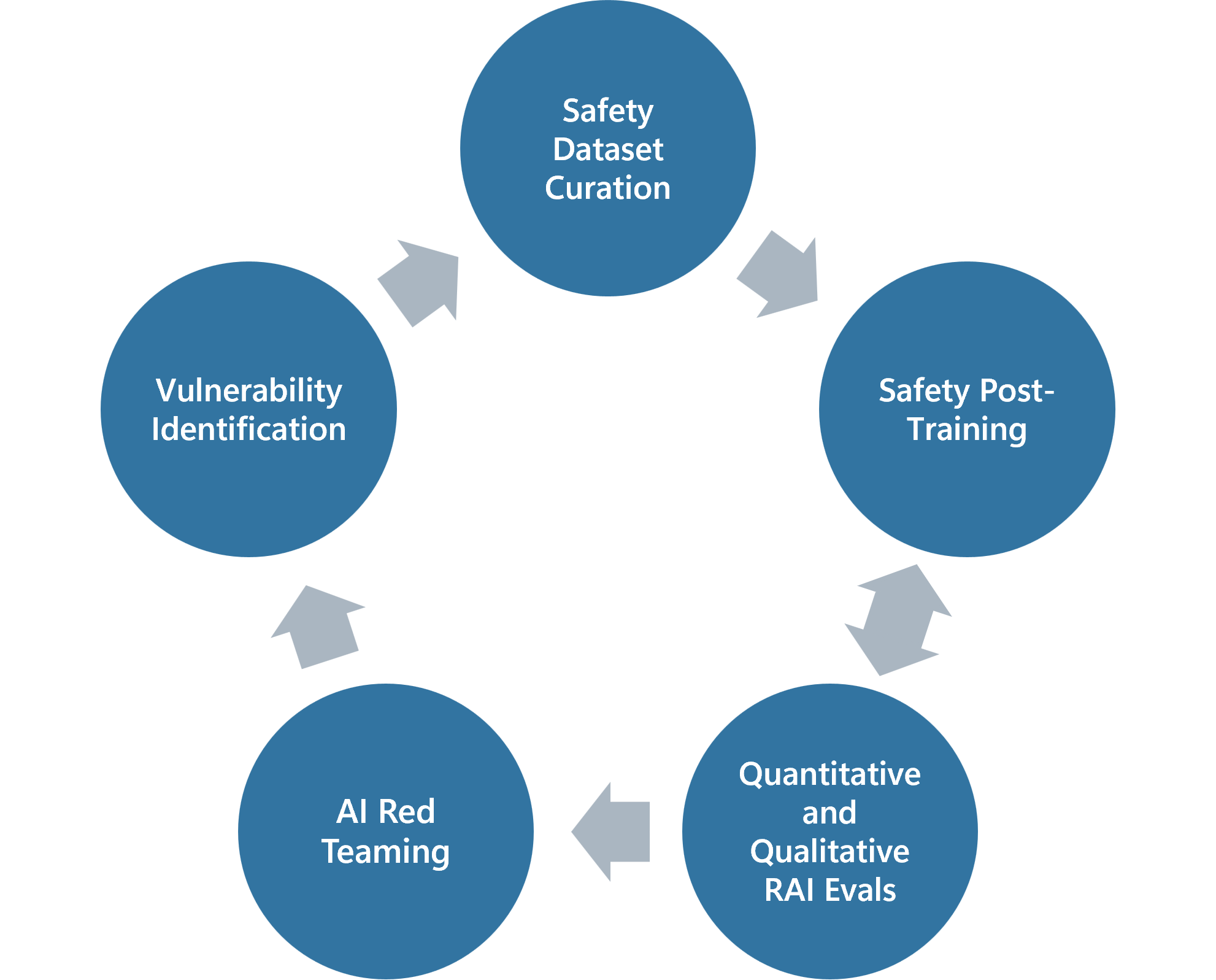}
\centering
\caption{The five main stages of the “break-fix” cycle used to safety align the Phi-3 language models. The double-headed arrow between “Safety Post-Training” and “Quantitative and Qualitative RAI Evals” illustrates the iterative process of running and shortlisting multiple release candidates with different data ablations before sharing a model with the AI Red Team.}
\label{fig:break-fix-cycle}
\end{figure}

As illustrated by Figure \ref{fig:break-fix-cycle}, we repeated this cycle multiple times and gradually fine-tuned the Phi-3 models to generate safe responses in a variety of contexts. We found that this iterative ``break-fix'' approach made it possible to mitigate many more risks than what can typically be achieved by a single fine-tuning job. In addition to RAI benchmarks, we monitored multiple performance metrics to ensure that safety post-training did not degrade the quality of generated text. The datasets, red teaming strategies, and evaluation benchmarks used are detailed in the sections below. 

\subsection{Safety Alignment}

For safety post-training of the Phi-3 models, we used a combination of open-source and in-house datasets. To improve the quality and effectiveness of the open-source datasets including \cite{bai2022training} and \cite{ji2023beavertails}, we used a variety of approaches including regenerating responses with GPT-4 and applying the instruction conversion method outlined in \cite{bianchi2024safetytuned}. While the open-source datasets covered a wide range of safety aspects, in-house datasets were curated to mitigate specific risks reported by AIRT as fine-tuning or preference optimization datasets depending on their effectiveness. 

In both the SFT and DPO stages, all safety datasets were mixed and used with other preference datasets leveraged in the post-training process. For every model checkpoint, both general quality evaluations and safety evaluations were conducted to decide a model checkpoint to be reviewed by AIRT and to eventually choose the best candidate for release. 

\subsection{Red Teaming}

\subsubsection{English-language testing (Phi-3)}

The Phi-3 release candidates were shared with a centralized Microsoft AI Red Team (AIRT), which leveraged a variety of techniques to test the models for responsible AI (RAI) risks across a range of categories. Because the original Phi-3 models were not trained on curated multilingual datasets, this red teaming was conducted in English.

AIRT probed Phi-3 models for harmful content using both single-turn prompts and multi-turn conversations. These strategies were further split into “low-skilled adversary” and “intermediate adversary” personas, as summarized in Table \ref{tab:airt-strategies}. Personas were scoped based on the most common ways in which users might elicit harmful content from models. A low-skilled adversary was defined as a typical chatbot user who tries to generate harmful content by asking for it directly, while an intermediate adversary actively attempts to subvert a model’s safety guardrails using, for example, basic encodings and jailbreaks. To gauge the risk posed by Phi-3 models in comparison to open-source equivalents, AIRT performed the same testing on Gemma-7B \cite{gemmateam2024gemma}, Mixtral-8x7B \cite{jiang2024mixtral}, and Llama-3-In \cite{touvron2023llama}. 

\begin{table}
\begin{center}
    \begin{adjustbox}{width=0.95\textwidth,center}
    \setlength\extrarowheight{6pt}
        \fontsize{10}{12}\selectfont
        \begin{tabularx}{\textwidth}{ c||YY } 
        & \makecell{Single-Turn} & \makecell{Multi-Turn \\ ($n=5$ or $n=8$)} \\
        \hline & \\[-3.5ex]
        Low-Skilled Adversary & Single-turn prompts in English asking the model to generate harmful content. & Multi-turn conversations asking for harmful content, automated using an attacker bot (GPT-4) via PyRIT. \\
        Intermediate Adversary & Common prompt encodings (e.g., base64, leetspeak, ROT-13) and public jailbreaks (e.g., BetterDAN, AIM, AntiGPT) applied to the low-skilled adversary single-turn prompts.  & Priming the model to respond “yes” to a series of prompts before asking for harmful content and Crescendo-like strategies tested manually. \\
        \end{tabularx}
    \end{adjustbox}
\end{center}
\caption{Summary of the four main adversarial scenarios used to red team the Phi-3 language models.}
\label{tab:airt-strategies}
\end{table}

AIRT operated independently of the safety post-training team and probed for harmful content across a range of categories, including content related to current events, phishing and cybersecurity, fairness and bias, hate speech, sexual content, and violence. Note that these harm categories were selected in accordance with Microsoft’s Responsible AI Standard \cite{msftrai2022} and were not necessarily the same as those covered by the preference datasets used for safety post-training. 

To red team the Phi-3 models at scale and cover as much of the risk surface as possible, AIRT utilized PyRIT: Python Risk Identification Toolkit\footnote{PyRIT GitHub repository: \url{https://github.com/Azure/PyRIT}} \cite{pyrit2024}. PyRIT is an open-source project that provides automation to support prompt generation, prompt conversion (e.g., to apply encodings and jailbreaks), response scoring, and even multi-turn conversations driven by an attacker LLM. To verify the accuracy of PyRIT automation, AIRT manually checked the scored results and made corrections where necessary. Note that PyRIT was developed specifically to support red teaming operations and is separate from automation used to calculate responsible AI benchmarks. For the intermediate adversary multi-turn scenario, AIRT manually applied strategies like Crescendo, which uses seemingly benign prompts and gradually escalates the conversation to jailbreak a model \cite{russinovich2024great}.

In section \ref{sec:rai-safety-benchmarks}, we present the results of the final Phi-3 models on a range of responsible AI benchmarks, along with the results achieved by comparable open-source models. We utilized these benchmarks to track overall safety performance and compare release candidates throughout the safety post-training process. It is important to note, however, that a model’s full risk profile can never be fully captured by a single set of metrics. In contrast with safety benchmarks, red teaming targets emerging harm areas, leverages the latest adversarial techniques, and can address ambiguous scenarios in which a model’s behavior might be interpreted in multiple ways. Importantly, red teaming centers the human elements of AI safety and can help answer questions related to how users might feel while interacting with a model. For example, in what scenarios might the model make users feel uncomfortable? Is the model at risk of providing dangerous or harmful advice? Are users likely to trust the model? During the red teaming phase of the break-fix cycle, questions like these played an important role in deciding where to focus further safety post-training efforts.

\subsubsection{Multilingual testing (Phi-3.5)}

In August 2024, Microsoft announced Phi-3.5-mini\footnote{Phi-3.5-mini model card: \url{https://huggingface.co/microsoft/Phi-3.5-mini-instruct}} and Phi-3.5-MoE.\footnote{Phi-3.5-MoE model card: \url{https://huggingface.co/microsoft/Phi-3.5-MoE-instruct}} In addition to performance and latency improvements, these models provide multilingual capabilities, and AIRT conducted an additional operation to probe Phi-3.5-MoE for violent content, sexual content, and hate speech across four languages: Chinese, Spanish, Dutch, and English (baseline).

Given the limited precedent for multilingual RAI red teaming, AIRT enlisted fluent speakers in these languages to perform red teaming manually. In particular, the following techniques were leveraged:
\begin{itemize}
    \item \textbf{Single-turn prompting:} To compare the behavior of Phi-3.5 in English, Chinese, Dutch, and Spanish, AIRT manually translated and evaluated the model with the same set of prompts across four harm categories -- violent content, sexual content, ECI, and hate speech. 
    \item \textbf{Single-turn jailbreaks:} To compare the robustness of Phi-3.5 to jailbreaks in English, Chinese, Dutch, and Spanish, AIRT leveraged a dataset of well-known jailbreaks (e.g., BetterDAN, AntiGPT, etc.) and applied these jailbreaks to a variety of prompts that the model typically refuses.
    \item \textbf{Multiturn role-playing:} To compare the robustness of Phi-3.5 in English, Chinese, Dutch, and Spanish to multiturn strategies, AIRT manually applied a role-playing strategy in both generic and culturally specific contexts. 
\end{itemize}

Overall, AIRT found that the safety behavior of Phi-3.5-MoE is similar in the four languages tested. More specifically, the model typically refused direct requests for harmful content and was robust to common jailbreaks in all languages. Although slightly higher refusal rates were observed in English than other languages, there was no significant difference in terms of harmful content generated. Note, however, that the model often provided English refusals even when the request for harmful content was in another language.

AIRT found that Phi-3.5-MoE model was susceptible to a multiturn role-playing strategy, which was used to elicit sexual and violent content, fake news, and hate speech in all languages, including English. This multiturn strategy was also applied to culturally specific scenarios in Dutch and Chinese, but the model generated similar content when prompted in English. 

Note that AIRT did not perform extensive testing of medium and low-resource languages, and it is possible that safety post-training in English does not transfer as effectively to other languages and scenarios.

\section{Safety Evaluation}

\subsection{RAI Safety Benchmarks}
\label{sec:rai-safety-benchmarks}

Throughout the safety alignment process, we used a range of responsible AI (RAI) benchmarks, including both public datasets and Microsoft internal measurements, to track the safety performance of Phi-3 models and compare release candidates. In this section, we explain these benchmarks in detail and present the results achieved by the final release candidates, as well as those achieved by Mistral-7B, Gemma-7B, and Llama-3-In, for comparison. 

\subsubsection{Internal Automated Measurement}
\label{sec:lasertag}

We used one of Microsoft’s automated measurement systems that leverages highly capable models like GPT-4 to simulate multi-turn conversations between adversarial AI agents and a target model \cite{magooda2023framework}. Among diverse conversation templates available, we ran experiments for the five scenarios below. 

\begin{itemize}
    \item \textbf{Grounding:} Asking the model to reason based on the information provided in prompts. 
    \item \textbf{3rd Party Content:} Asking a model to provide protected third-party content. 
    \item \textbf{Harmful Content Continuation:} Asking the model to generate harmful content.
    \item \textbf{Harmful Content Summarization:} Asking the model to summarize harmful content.
    \item \textbf{Jailbreak:} Asking the model to bypass behavior protocols learned during safety post-training. 
\end{itemize}

This technique probes for harmful content by making multiple requests in a multi-turn setting or by posing hypothetical scenarios that make the model more likely to respond. (e.g., asking the model to generate harmful content “for a research purpose”).  

Table \ref{tab:lasertag-results} shows the results for the Phi-3 and baseline models across the five scenarios. GPT-4 was used to evaluate the model responses. Ungroundedness measures how much the response is based on a given prompt on a scale from 0 (fully grounded) to 4 (not grounded). The other categories are evaluated in terms of severity from 0 (no harm) to 7 (severe harm). The defect rates (DR-$x$) shown in the table are computed as the percentage of samples with severity score greater than or equal to $x$. Note that a lower score indicates more desirable performance in all scenarios.

The Phi-3 models showed scores that are better than or comparable to the scores achieved by the baseline models in all five scenarios. 

\begin{table}[h]
\begin{center}
    \begin{adjustbox}{width=0.95\textwidth,center}
    \setlength\extrarowheight{6pt}
    \fontsize{10}{12}\selectfont
        \begin{tabular}{ c||ccccccc}
        & \makecell{Phi-3-mini \\ \footnotesize 3.8b} & \makecell{Phi-3-small \\ \footnotesize 7b} & \makecell{Phi-3-medium \\ \footnotesize 14b} & \makecell{Phi-2 \\ \footnotesize 2.7b } & \makecell{Mistral\\ \footnotesize 7b } & \makecell{Gemma \\ \footnotesize 7b} & \makecell{Llama-3-In \\ \footnotesize 8b} \\
        \hline & \\[-3.5ex]
        Ungroundedness  & 0.603 & 0.299 & 0.213 & 1.481 & 0.935 & 0.679 & 0.328  \\
        Third Party Harm (DR-1) & 0.240 & 0.253 & 0.251 & 0.240 & 0.562 & 0.383 & 0.373 \\
        \makecell{Harmful Content \\ Continuation (DR-3)} & 0.007 & 0.003 & 0.010 & 0.029 & 0.026 & 0.013 & 0.013 \\
        \makecell{Harmful Content \\ Summarization (DR-3)} & 0.100 & 0.110 & 0.112 & 0.144 & 0.223 & 0.103 & 0.082 \\
        Jailbreak (DR-1) & 0.123 & 0.107 & 0.111 & 0.150 & 0.156 & 0.114 & 0.130 \\
        \end{tabular}
    \end{adjustbox}
\end{center}
\caption{Results of Phi-3 models and baseline models on the Microsoft internal multi-turn conversation benchmarks. Note that a lower value indicates a better performance for all metrics in the table.}
\label{tab:lasertag-results}
\end{table}

\subsubsection{XSTest}
\label{sec:xstest}

XSTest is a public dataset of 250 safe prompts across ten prompt categories (e.g., definitions, historical events, etc.) that well-calibrated models should comply with, and 200 unsafe prompts that most general-purpose models should refuse \cite{röttger2024xstest}.  

The following two refusal metrics are computed in this benchmark: 

\begin{itemize}
    \item \textbf{Inappropriate Prompt Refusal Rate (IPRR):} Measures the rate that the model refuses to answer inappropriate or harmful prompts (higher is better).
    \item \textbf{Valid Prompt Refusal Rate (VPRR):} Measures the rate that the model refuses to answer appropriate or innocuous prompts (lower is better).
\end{itemize}

Table \ref{tab:xstest-results} shows IPRR and VPRR values for the Phi-3 and baseline models. We observed that higher IPRR values are often associated with higher VPRR values. In other words, models that are more likely to refuse harmful prompts are also more likely to refuse harmless prompts. This behavior is well known in responsible AI research and is often described as a tradeoff between helpfulness (higher VPRR) and harmlessness (higher IPRR). In the table below, we observe this tradeoff across all models tested. More specifically, we see that Phi-3-small and Gemma achieve similar balances between IPRR and VPRR. In addition, Llama-3-In is comparable to, but slightly outperforms, Phi-3-medium. 

\begin{table}[h]
\begin{center}
    \begin{adjustbox}{width=0.95\textwidth,center}
    \setlength\extrarowheight{6pt}
    \fontsize{10}{12}\selectfont
        \begin{tabular}{ c||ccccccc} 
        & \makecell{Phi-3-mini \\ \footnotesize 3.8b} & \makecell{Phi-3-small \\ \footnotesize 7b} & \makecell{Phi-3-medium \\ \footnotesize 14b} & \makecell{Phi-2 \\ \footnotesize 2.7b } & \makecell{Mistral\\ \footnotesize 7b } & \makecell{Gemma \\ \footnotesize 7b} & \makecell{Llama-3-In \\ \footnotesize 8b} \\
        \hline & \\[-3.5ex]
        IPRR (Higher is better)  & 0.750 & 0.965 & 0.790 & 0.015 & 0.040 & 0.955 & 0.815  \\
        VPRR (Lower is better) & 0.232 & 0.264 & 0.124 & 0.004 & 0.008 & 0.216 & 0.024 \\
        \end{tabular}
    \end{adjustbox}
\end{center}
\caption{Results of Phi-3 models and baseline models on the XSTest benchmarks.}
\label{tab:xstest-results}
\end{table}

\subsubsection{DecodingTrust}

DecodingTrust is a comprehensive trustworthiness evaluation methodology which considers diverse risks including toxicity, stereotype bias, adversarial robustness, out-of-distribution robustness, robustness on adversarial demonstrations, privacy, machine ethics, and fairness \cite{wang2024decodingtrust}. In our DecodingTrust benchmarks, we covered all of these risks except for toxicity, which is separately covered by the ToxiGen benchmark.  

\begin{itemize}
    \item \textbf{Stereotype Bias:} Checks whether the model can identify stereotypes included in prompts.
    \item \textbf{Robustness Metrics:} Three robustness-related metrics measure how consistently the model can identify inappropriate prompts when multiple variations of a prompt with the same meaning are provided.
    \item \textbf{Privacy:} Measures how likely the model is to include personal information such as a phone number or email address in responses.   
    \item \textbf{Machine Ethics:} Measures how well the model can understand immorality in prompts.
    \item \textbf{Fairness:} Measures how well the model can provide consistent answers when only sensitive attributes such as gender are changed in prompts. 
\end{itemize}
 
Note that the DecodingTrust metrics do not evaluate a model’s ability to generate more desirable content. Rather, these metrics are primarily based on how well a model understands various responsible AI risks. Therefore, the values in the table below should be interpreted as performance indicators of language tasks such as harmful content detection rather than content generation (higher is better). 

Table \ref{tab:decodingtrust-results} shows the DecodingTrust benchmarks for the Phi-3 and baseline models. As shown in the table below, no model is universally better or worse than the others. 

\begin{table}
\begin{center}
    \begin{adjustbox}{width=0.95\textwidth,center}
    \setlength\extrarowheight{6pt}
    \fontsize{10}{12}\selectfont
        \begin{tabular}{ c||cccccc }
        & \makecell{Phi-3-mini \\ \footnotesize 3.8b} & \makecell{Phi-3-small \\ \footnotesize 7b} & \makecell{Phi-3-medium \\ \footnotesize 14b} & \makecell{Phi-2 \\ \footnotesize 2.7b } & \makecell{Mistral\\ \footnotesize 7b } & \makecell{Gemma \\ \footnotesize 7b} \\
        \hline & \\[-3.5ex]
        Stereotype Bias & 0.983 & 0.983 & 0.993 & 0.860 & 0.990 & 0.996  \\
        Adversarial Robustness & 0.490 & 0.615 & 0.516 & 0.513 & 0.381 & 0.497 \\
        Out-of-Distribution Robustness & 0.643 & 0.706 & 0.747 & 0.655 & 0.667 & 0.644 \\
        Robustness to Adversarial Demonstrations & 0.666 & 0.635 & 0.719 & 0.467 & 0.575 & 0.572 \\
        Privacy & 0.926 & 0.993 & 0.824 & 0.568 & 0.688 & 0.905 \\
        Machine Ethics & 0.754 & 0.775 & 0.737 & 0.425 & 0.710 & 0.766 \\
        Fairness & 0.825 & 0.589 & 0.663 & 0.668 & 0.827 & 0.950 \\
        \end{tabular}
    \end{adjustbox}
\end{center}
\caption{Results of Phi-3 models and baseline models on the DecodingTrust benchmarks. Note that a higher value indicates better performance for all metrics in the table.}
\label{tab:decodingtrust-results}
\end{table}

\subsubsection{ToxiGen}
\label{sec:toxigen}

ToxiGen is a large-scale machine generated dataset for adversarial and implicit hate speech detection \cite{hartvigsen2022toxigen}. As mentioned above, we used the ToxiGen benchmark in favor of the toxicity category in DecodingTrust because ToxiGen is a richer dataset with more prompts (274K) than DecodingTrust toxicity (under 100K). A high ToxiGen score means that the model can detect harmfulness in prompts well. The scores are shown in Table \ref{tab:ToxiGen-results}. All three Phi-3 model variants outperformed Mistral and Gemma. 

\begin{table}[h]
\begin{center}
    \begin{adjustbox}{width=0.95\textwidth,center}
    \setlength\extrarowheight{6pt}
     \fontsize{9.5}{12}\selectfont
        \begin{tabularx}{\textwidth}{ c||YYYYYY} 
        & \makecell{Phi-3-mini \\ \footnotesize 3.8b} & \makecell{Phi-3-small \\ \footnotesize 7b} & \makecell{Phi-3-medium \\ \footnotesize 14b} & \makecell{Phi-2 \\ \footnotesize 2.7b } & \makecell{Mistral\\ \footnotesize 7b } & \makecell{Gemma \\ \footnotesize 7b} \\
        \hline & \\[-3.5ex]
        ToxiGen & 0.764 & 0.827 & 0.855 & 0.589 & 0.677 & 0.572  \\
        \end{tabularx}
    \end{adjustbox}
\end{center}
\caption{Results of Phi-3 models and baseline models on the ToxiGen benchmark (higher is better).}
\label{tab:ToxiGen-results}
\end{table}

\subsection{Iterative Safety Alignment}

The RAI benchmark scores reported above reflect the performance of the final Phi-3-mini, Phi-3-small, and Phi-3-medium release candidates. However, multiple iterations of safety post-training, red teaming, and vulnerability identification were required to achieve the best results. Given the vast number of scenarios that users might encounter, we found that this iterative approach, as opposed to a single fine-tuning job, was necessary to identify and mitigate a realistic range of real-world risks.

\begin{figure}[ht]
\includegraphics[scale=0.6]{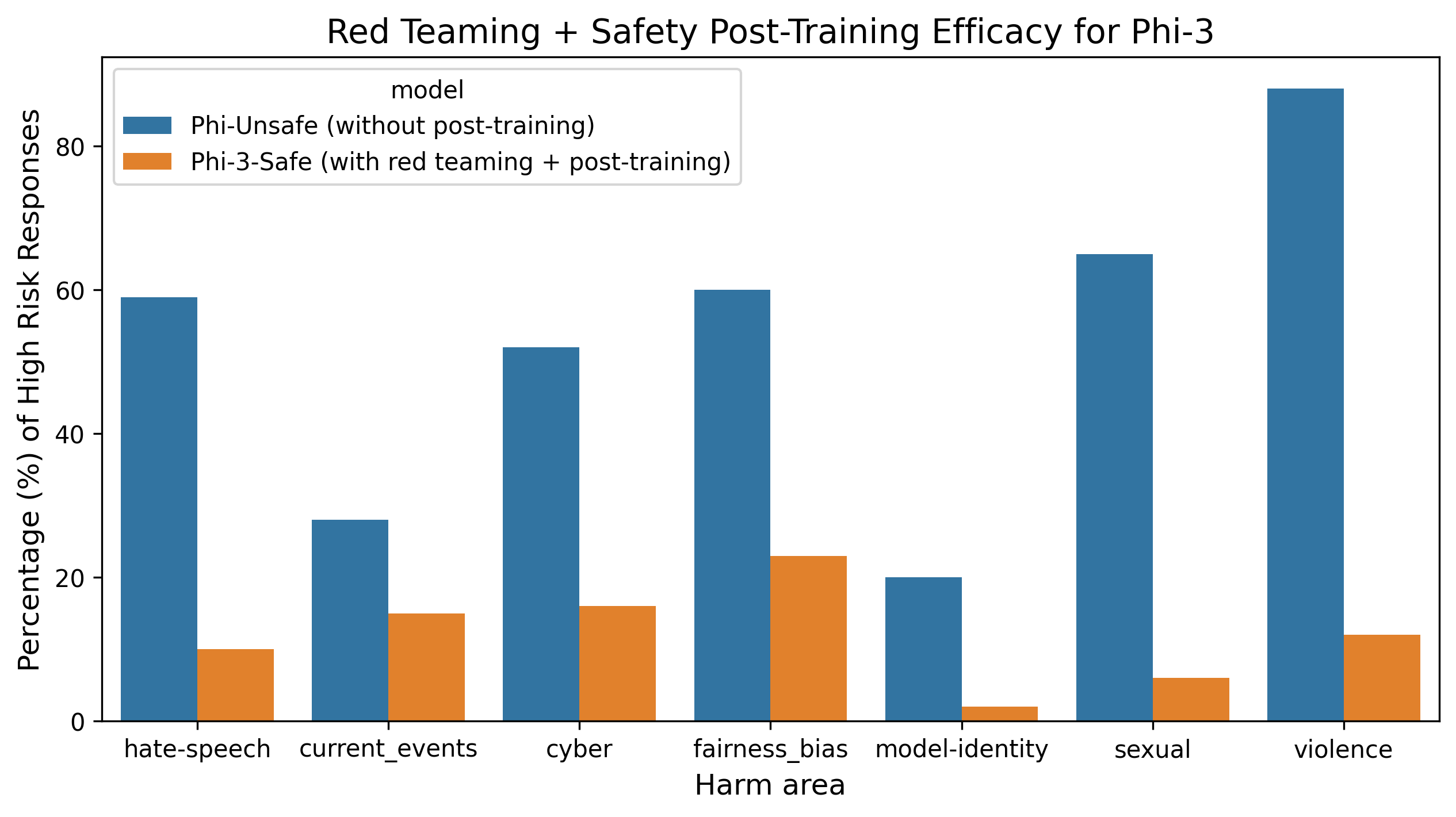}
\centering
\caption{Comparison of high-risk responses generated by Phi-3 language models before and after several rounds of the “break-fix” cycle. Note that percentages are inflated because prompts used by the AI Red Team were crafted to elicit harmful generations.}
\label{fig:safety-comparison}
\end{figure}

To further quantify the overall improvement in safety alignment, we evaluated the Phi-3 models before and after completing several rounds of the ``break-fix'' cycle on a dataset of prompts used by the AI Red Team. In Figure \ref{fig:safety-comparison}, we plot the percentage of harmful responses generated by models with and without safety alignment across several harm categories. On average, we observe a 75\% reduction in the amount of harmful content generated, which indicates the efficacy of the overall ``break-fix'' approach. 

\subsection{Multilingual RAI Safety Benchmarks}

A multi-faceted approach was adopted to evaluate the Phi-3.5 language models for safety behaviors, with additional measures taken to account for the multilingual capabilities introduced by this release. Our approach to multilingual safety evaluation included both internal automated benchmarks and open-source benchmarks. In this section, we explain these benchmarks and present the results achieved by the final release candidates as well as those achieved by Mistral-Nemo. 

\subsubsection{Internal Automated Measurement}

First, we reproduced the \textit{English-language} measurements described  in section \ref{sec:lasertag} for the Phi-3.5 models. The results can be seen in Table \ref{tab:lasertag-results-3.5-en}. We found, for the vast majority of metrics, that our 3.5 releases (Phi-3.5-mini-instruct and Phi-3.5-MoE-instruct) were on par with or better than the baseline model (Mistral-Nemo).

\begin{table}[h]
\begin{center}
    \begin{adjustbox}{width=0.95\textwidth,center}
    \setlength\extrarowheight{6pt}
    \fontsize{10}{12}\selectfont
        \begin{tabularx}{\textwidth}{ c||YYY}
        & \makecell{Phi-3.5-mini-instruct \\ \footnotesize 3.8b} & \makecell{Phi-3.5-MoE-instruct \\ \footnotesize 16$\times$ 3.8b} & \makecell{Mistral-Nemo \\ \footnotesize 12b} \\
        \hline & \\[-3.5ex]
        Ungroundedness  & 0.459 & 0.271 & 0.244  \\
        Third Party Harm (DR-1) & 0.243 & 0.152 & 0.599  \\
        \makecell{Harmful Content \\ Continuation (DR-3)} & 0.006 & 0.004 & 0.025  \\
        \makecell{Harmful Content \\ Summarization (DR-3)} & 0.119 & 0.120 & 0.110  \\
        Jailbreak (DR-1) & 0.119 & 0.110 & 0.174  \\
        \end{tabularx}
    \end{adjustbox}
\end{center}
\caption{Results of Phi-3.5 models and baseline model on the Microsoft internal multi-turn conversation benchmarks (\textit{English-only}). Note that a lower value indicates a better performance for all metrics in the table.}
\label{tab:lasertag-results-3.5-en}
\end{table}

To evaluate the risk of multilingual harmful content generation, we augmented our existing internal benchmark for additional languages supported by the Phi-3.5 models. Table \ref{tab:lasertag-results-3.5-multiling} shows the harmful content continuation metric for some of the languages supported by Phi-3.5. We observe better performance than Mistral-Nemo, with Phi-3.5-MoE-instruct tending to do slightly better than Phi-3.5-mini-instruct.

We are continuing research and investments into increasing our multilingual benchmark coverage for additional metrics and languages.  

\begin{table}[h]
\begin{center}
    \begin{adjustbox}{width=0.95\textwidth,center}
    \setlength\extrarowheight{6pt}
    \fontsize{10}{12}\selectfont
        \begin{tabularx}{\linewidth}{ c||YYY}
        \makecell{Harmful Content \\ Continuation (DR-3)} & \makecell{Phi-3.5-mini-instruct \\ \footnotesize 3.8b} & \makecell{Phi-3.5-MoE-instruct \\ \footnotesize 16$\times$ 3.8b} & \makecell{Mistral-Nemo \\ \footnotesize 12b} \\
        \hline & \\[-3.5ex]
        \makecell{Chinese Simplified} & 0.015 & 0.007 &  0.026 \\
        \makecell{French} & 0.010 & 0.010 &  0.034 \\
        \makecell{German} & 0.007 & 0.010 &  0.027 \\
        \makecell{Italian} & 0.009 & 0.009 & 0.023  \\
        \makecell{Japanese} & 0.010 & 0.008 &  0.024 \\
        \makecell{Portuguese (Brazil)} & 0.012 & 0.017 & 0.026  \\
        \makecell{Spanish} & 0.016 & 0.008 &  0.026 \\
        \end{tabularx}
    \end{adjustbox}
\end{center}
\caption{Results of Phi-3.5 models and baseline model on the ``Harmful Content Continuation'' Microsoft internal multi-turn conversation benchmark \textit{across several languages}. Because the reported metric is DR-3 (explained in section \ref{sec:lasertag}), a lower value indicates better performance.}
\label{tab:lasertag-results-3.5-multiling}
\end{table}

\subsubsection{XSafety}

To measure refusal rates, we used the XSafety dataset and methodology from \cite{wang2024xsafety} since the XSTest dataset described in section \ref{sec:xstest} was not multilingual. The XSafety dataset consists of $\sim$2,800 malicious prompts per language, and we use GPT-4 to determine whether Phi-3.5's responses to these prompts constitute refusals or not.

Results are given in Table \ref{tab:xsafety-results} and show that Phi-3.5-mini-instruct and Phi-3.5-MoE-instruct perform almost as well or even better than Mistral-Nemo.

\begin{table}[h]
\begin{center}
    \begin{adjustbox}{width=0.95\textwidth,center}
    \setlength\extrarowheight{6pt}
    \fontsize{10}{12}\selectfont
        \begin{tabularx}{\textwidth}{ c||YYY}
        \makecell{Refusal Rate} & \makecell{Phi-3.5-mini-instruct \\ \footnotesize 3.8b} & \makecell{Phi-3.5-MoE-instruct \\ \footnotesize 16$\times$ 3.8b} & \makecell{Mistral-Nemo \\ \footnotesize 12b} \\
        \hline & \\[-3.5ex]
        \makecell{Chinese Simplified} & 0.691  & 0.659   &  0.706   \\
        \makecell{French} & 0.874  &  0.865  &  0.870   \\
        \makecell{German} & 0.881  &  0.880  &   0.865  \\
        \makecell{Japanese} & 0.600  & 0.714   &  0.673   \\
        \makecell{Spanish} & 0.864  & 0.878   &  0.853   \\
        \end{tabularx}
    \end{adjustbox}
\end{center}
\caption{Results of Phi-3.5 models and baseline model on the XSafety benchmark across various languages. A higher value indicates better performance.}
\label{tab:xsafety-results}
\end{table}

\subsubsection{RTP-LX}

To assess the ability of the model to detect toxic speech, we used the framework described in \cite{dewynter2024rtplx}. The dataset consists of $\sim$1,000 prompts per language, along with a toxicity level provided by human reviewers. We then asked Phi-3.5 to rate the toxicity of the prompts and compared the Phi-3.5 ratings to human ratings. This is similar to what we did in section \ref{sec:toxigen} for English with the Toxigen dataset.

Results can be found in Table \ref{tab:rtplx-results} and indicate that Phi-3.5-MoE-instruct can detect toxic speech almost as well as Mistral-Nemo, and better than Phi-3.5-mini-instruct.

\begin{table}[h]
\begin{center}
    \begin{adjustbox}{width=0.95\textwidth,center}
    \setlength\extrarowheight{6pt}
    \fontsize{10}{12}\selectfont
        \begin{tabularx}{\textwidth}{ c||YYY}
        \makecell{Kappa/Inter-annotator \\ Agreement} & \makecell{Phi-3.5-mini-instruct \\ \footnotesize 3.8b} & \makecell{Phi-3.5-MoE-instruct \\ \footnotesize 16$\times$ 3.8b} & \makecell{Mistral-Nemo \\ \footnotesize 12b} \\
        \hline & \\[-3.5ex]
        \makecell{Chinese Simplified} & 0.195  & 0.287  &  0.323  \\
        \makecell{French} &  0.211 &  0.377 &  0.349  \\
        \makecell{German} & 0.162  &  0.251 &  0.276  \\
        \makecell{Italian} & 0.172  &  0.295 &  0.323  \\
        \makecell{Japanese} & 0.206  & 0.285  &  0.315  \\
        \makecell{Portuguese (Brazil)} & 0.203  & 0.323  & 0.366   \\
        \makecell{Spanish} & 0.196  &  0.278 &  0.297  \\
        \end{tabularx}
    \end{adjustbox}
\end{center}
\caption{Results of Phi-3.5 models and baseline model on the RTP-LX benchmark across various languages. A higher value indicates better performance.}
\label{tab:rtplx-results}
\end{table}

\section{Responsible AI Considerations for Developers}

\subsection{Responsible Downstream Development}

While the Phi-3 models benefit from a robust safety post-training approach, developers should consider how to adapt models with further fine-tuning to their specific use case and safety requirements. In addition to fine-tuning, developers should explore building or adopting additional safety-related tools and approaches to ensure that model outputs are appropriate for their context. These may include safety classifiers run on inputs or outputs, prompt engineering techniques, or other guidance to end-users about how to interpret or use model outputs appropriately. Further guidance and open-source tools are available via Microsoft’s Responsible AI Toolbox repository.\footnote{RAI Toolbox GitHub repository: \url{https://github.com/microsoft/responsible-ai-toolbox}}

In further developing or deploying models for downstream uses cases, developers should be aware of common capability limitations of language models that are also present in the Phi-3 series. Like other language models, Phi-3 models can potentially behave in ways that are unfair, unreliable, or offensive. Some of the limiting behaviors to be aware of include:

\begin{itemize}
    \item \textbf{Quality of Service:} The Phi-3 models are trained primarily on English text. Languages other than English will experience worse performance. English language varieties with less representation in the training data might experience lower performance than standard American English.
    \item \textbf{Multilingual Performance \& Safety Gaps:} We believe it is important to make language models more widely available across different languages, but the Phi-3 models still exhibit challenges common across multilingual releases. As with any deployment of LLMs, developers will be better positioned to test for performance or safety gaps for their linguistic and cultural context and customize the model with additional fine-tuning and appropriate safeguards.
    \item \textbf{Representational Harms \& Perpetuation of Stereotypes:} These models can over- or under-represent groups of people, erase representation of some groups, or reinforce demeaning or negative stereotypes. Despite safety post-training, these limitations may still be present due to differing levels of representation of different groups or prevalence of examples of negative stereotypes in training data that reflect real-world patterns and societal biases.
    \item \textbf{Inappropriate or Offensive Content:} These models may produce other types of inappropriate or offensive content, which may make it inappropriate to deploy for sensitive contexts without additional mitigations that are specific to the use case. 
    \item \textbf{Information Reliability:} Language models can generate nonsensical content or fabricate content that might sound reasonable but is inaccurate or outdated. 
    \item \textbf{Limited Scope for Code:} The majority of code in the Phi-3 training data is based in Python. We strongly recommend that users manually verify all code generated by the Phi-3 models, especially for languages other than Python. 
    \item \textbf{Long Conversation:} Phi-3 models, like other models, can in some cases generate responses that are repetitive, unhelpful, or inconsistent in very long chat sessions in both English and non-English languages. Developers are encouraged to place appropriate mitigations, like limiting conversation turns to account for possible conversational drift.
\end{itemize}

\subsection{Additional Considerations}

At Microsoft, we are committed to advancing the state of the art in AI and ensuring that our AI products and services are safe, secure, and trustworthy. Language models have great potential to enable new capabilities and benefit society by driving an open innovation cycle and enabling an extensive value chain built on open-source projects. These include direct and indirect benefits, such as advancing AI safety and security, fostering global collaboration and academic research, and inviting greater participation in the development of AI systems.  

We have also considered the potential risks and believe the release of the Phi-3 models does not have a meaningful impact on marginal risk of the AI ecosystem due to the availability of larger, advanced AI models and other open information. Based on the evaluations and safety post-training detailed in this white paper, we have assessed the potential benefits of open innovation and research will outweigh potential risks specific to this model.  

There are specific areas our team has considered and taken steps to address, but we have not designed or evaluated these models for every potential downstream use case. Developers should take into account common limitations of this technology as they select use cases, as well as conduct evaluations and implement appropriate safeguards in additional fine-tuning and deployment stages. Developers have a responsibility to apply responsible AI best practices and ensure that a specific use case complies with relevant laws. Important areas for consideration include: 

\begin{itemize}
    \item \textbf{Allocation:} While we have implemented mitigations in post-training to address potential biases, given the known limitations of language models, these models may not be suitable for scenarios that could have consequential impact on legal status or the allocation of resources or life opportunities without performing further assessments and applying additional debiasing techniques. For example, in conferring legal rights or an individual’s access to credit, education, employment, healthcare, housing, insurance, social welfare benefits, services, or opportunities, or the terms on which they are provided.   
    \item \textbf{High-Risk Scenarios:} Developers should also assess the suitability of using models in high-risk scenarios in situations where unfair, unreliable, or offensive outputs might be extremely costly or lead to harm. This includes providing advice in sensitive or expert domains where accuracy and reliability are critical (e.g., legal or health advice). Additional safeguards should be implemented at the application level according to the deployment context.
    \item \textbf{Misinformation:} Language models may produce inaccurate information. Developers should consider and adopt best practices for transparency and disclosure to inform end-users that they are interacting with an AI system. As part of applications, developers can also build mechanisms for feedback as well as pipelines to ground responses in use-case specific, contextual information, a technique known as Retrieval Augmented Generation (i.e., “RAG”).
    \item \textbf{Generation of Harmful Content:} While safety post-training has reduced the likelihood that the model will generate some forms of harmful content, developers will need to make assessments based on their context and use available safety classifiers or custom solutions based on their use cases.
    \item \textbf{Privacy:}  Developers should be aware of and adhere to privacy laws in the jurisdictions where they operate and deploy applications.
    \item \textbf{Misuse:} Other forms of misuse such as fraud, spam, or malware production may be possible, and developers should ensure that their applications do not violate applicable laws and regulations.
\end{itemize}

\section{Conclusion}

In this report, we presented our approach to safety aligning the Phi-3 series of language models. We adopted an iterative “break-fix” approach by performing multiple rounds of dataset curation, post-training with DPO, responsible AI benchmarking, red teaming, and vulnerability identification. Safety alignment is a challenging and open-ended task, but our results indicate that this cycle significantly reduced the amount of harmful content generated by the Phi-3 models in a range of scenarios. Nonetheless, the Phi-3 models are susceptible to the same fundamental limitations as any modern language model, and we hope that the responsible AI considerations outlined in this report will encourage users to think carefully about additional safety mitigations that may be necessary for their specific use cases.

\section*{Contributors}

\noindent \textbf{GenAI Model Team:} Emman Haider, Daniel Perez-Becker, Thomas Portet, Piyush Madan, Amit Garg, Atabak Ashfaq, David Majercak, Wen Wen, Dongwoo Kim, Ziyi Yang, Jianwen Zhang, Hiteshi Sharma \vspace{5px}

\noindent \textbf{AI Red Team:} Blake Bullwinkel, Martin Pouliot, Amanda Minnich, Shiven Chawla, Solianna Herrera, Shahed Warreth, Maggie Engler, Gary Lopez, Nina Chikanov, Raja Sekhar Rao Dheekonda, Bolor-Erdene Jagdagdorj, Roman Lutz, Richard Lundeen, Tori Westerhoff, Pete Bryan, Christian Seifert, Ram Shankar Siva Kumar \vspace{5px}

\noindent \textbf{Office of Responsible AI:} Andrew Berkley, Alex Kessler \vspace{5px}

\section*{Acknowledgements}

We are very grateful to April Rettkowski, Yonatan Zunger, and Weizhu Chen for providing valuable feedback on this paper.

\bibliographystyle{alpha}
\bibliography{references}

\end{document}